\documentclass[10pt, a4paper]{article}
\pdfoutput=1
\usepackage{lrec2022}
\usepackage{graphicx}
\usepackage{tabularx}
\usepackage{multirow}
\usepackage{booktabs}
\usepackage{paralist}
\usepackage{makecell}
\usepackage{xspace}
\newcommand{\F}{F$_1$\xspace}
\newcolumntype{R}{@{\extracolsep{5pt}}r@{\extracolsep{0pt}}}%
\newcolumntype{P}{@{\extracolsep{-2pt}}r@{\extracolsep{0pt}}}%

\usepackage{amsmath}
\usepackage{stackengine}
\usepackage[dvipsnames,table,xcdraw]{xcolor}
\newcommand{\surround}[1]{$\big[$#1$\big]$}
\newcommand{\stack}[2]{$\stackunder{\text{#1}}{\text{\tiny\sf #2}}$}

\newcommand{\medC}[1]{\surround{\textcolor{orange}{\stack{#1}{\textsc{medC}}}}}
\newcommand{\drug}[1]{\surround{\textcolor{Blue}{\stack{#1}{\textsc{drug}}}}}
\newcommand{\therapy}[1]{\surround{\textcolor{Blue}{\stack{#1}{\textsc{therapy}}}}}
\newcommand{\pathogen}[1]{\surround{\textcolor{ForestGreen}{\stack{#1}{\textsc{pathogen}}}}}
\newcommand{\substance}[1]{\surround{\textcolor{blue!70!black}{\stack{#1}{\textsc{substance}}}}}
\newcommand{\process}[1]{\surround{\textcolor{blue!70!black}{\stack{#1}{\textsc{process}}}}}
\newcommand{\diagnostics}[1]{\surround{\textcolor{Lavender}{\stack{#1}{\textsc{diagnostics}}}}}
\newcommand{\other}[1]{\surround{\textcolor{black}{\stack{#1}{\textsc{other}}}}}

\newcommand{\qol}[1]{\surround{\textcolor{SeaGreen}{\stack{#1}{\textsc{qol}}}}}
\newcommand{\socioEcon}[1]{\surround{\textcolor{Purple}{\stack{#1}{\textsc{socio-econ}}}}}

\newcommand{\dietary}[1]{\surround{\textcolor{Purple}{\stack{#1}{\textsc{dietary}}}}}
\newcommand{\habit}[1]{\surround{\textcolor{Purple}{\stack{#1}{\textsc{habitual}}}}}

\usepackage{tikz-dependency}

\usepackage{mparhack}

\usepackage{titlesec}
\titleformat{\section}{\normalfont\large\bfseries\center}{\thesection.}{1em}{}
\titleformat{\subsection}{\normalfont\SmallTitleFont\bfseries\raggedright}{\thesubsection.}{1em}{}
\titleformat{\subsubsection}{\normalfont\normalsize\bfseries\raggedright}{\thesubsubsection.}{1em}{}
\renewcommand\thesection{\arabic{section}}
\renewcommand\thesubsection{\thesection.\arabic{subsection}}
\renewcommand\thesubsubsection{\thesubsection.\arabic{subsubsection}}

\usepackage{epstopdf}
\usepackage[utf8]{inputenc}

\usepackage{hyperref}
\usepackage{xstring}

\usepackage{color}

\title{Recovering Patient Journeys:\\ A Corpus of Biomedical Entities and Relations on  Twitter (BEAR)}

\name{Amelie W\"uhrl, Roman Klinger} 

\address{Institut f\"ur Maschinelle Sprachverarbeitung, University of Stuttgart \\
  Pfaffenwaldring 5b, 70569 Stuttgart, Germany \\
  \{amelie.wuehrl, roman.klinger\}@ims.uni-stuttgart.de\\}

\abstract{ For a long time, text mining and information extraction for
  the medical domain has focused on scientific text generated by
  researchers. However, their direct access to individual patient
  experiences or patient-doctor interactions is sometimes limited.
  Information provided on social media, e.g.,\ by patients and their
  relatives, complements the knowledge available in scientific text.
  It reflects the patient's journey and their subjective perspective
  on the process of developing symptoms, being diagnosed and offered a
  treatment, being cured or learning to live with a medical
  condition. The value of this type of data is therefore twofold:
  Firstly, it offers direct access to people's perspectives. Secondly,
  it might cover information that is not available elsewhere,
  including self-treatment or self-diagnoses.
  Named entity recognition and relation extraction are methods to
  structure information that is available in unstructured
  text. However, existing medical social media corpora focused on a
  comparably small set of entities and relations and were focused on
  particular domains, rather than putting the patient into the center
  of analyses. With this paper we contribute a corpus with a rich
  set of annotation layers following the motivation to uncover and
  model patients' journeys and experiences in more detail. We label 14
  entity classes (incl.\ environmental factors, diagnostics,
  biochemical processes, patients' quality-of-life descriptions,
  pathogens, medical conditions, and treatments) and 20 relation
  classes (e.g.,\ prevents, influences, interactions, causes) most of
  which have not been considered before for social media data. The
  publicly available dataset consists of 2,100 tweets with
  $\approx$6,000 entity and $\approx$3,000 relation annotations. In a
  corpus analysis we
  find that over 80\,\% of documents contain relevant entities. Over
  50\,\% of tweets express relations which we consider essential for
  uncovering patients' narratives about their journeys.\\[\baselineskip]
}

\begin{document}

\maketitleabstract

\begin{figure}[b]
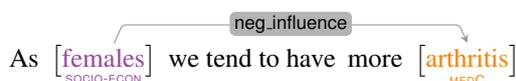

\centering

\begin{dependency}[edge style={Gray}, label style={gray!60, fill=gray!60,font=\sf,text=black}]
\begin{deptext}
As \& \socioEcon{females} \&we tend to have\& more \& \medC{arthritis}\\
\end{deptext}
\depedge[edge unit distance=0.5ex]{2}{5}{neg\_influence}
\end{dependency}
  
  \caption{Example of our annotation scheme. }
  \label{fig:introducotry-example}
\end{figure}

\section{Introduction}

On social media, doctors, patients, concerned relatives or other
laypeople frequently discuss medical information.  Twitter posts for
example contain opinions and recommendations about treatments, recounts of medical experiences, or hypotheses and assumptions about medical issues like in Figure~\ref{fig:introducotry-example}. This information is by design centered around
the patient. It is impacted by the patient's journey and their
subjective perspective on processes like developing symptoms, being
diagnosed and offered a treatment, being cured or learning to live
with a disease. This data offers direct access to people's
perspectives and covers information that is not available elsewhere,
e.g., aspects that might not be considered important or difficult
to assess in clinical settings. This includes, e.g., assessments of a
patient's quality of life (Table
\ref{tab:labeled-examples}, Ex.\ 2 and 5), or which environmental factors people
consider when talking about their health (Table
\ref{tab:labeled-examples}, Ex.\ 3 and 4).

At the same time, established resources and systems for text mining
and information extraction in the medical domain have mostly been
centered around scientific and biomedical text generated by
researchers. Such texts seldomly focus on individual patient's
experiences or patient-doctor interactions which makes the information
and knowledge contained in the text distant by nature. While
scientific resources contain high quality information, many studies
struggle with gender biases and population imbalance
\cite{weber-et-la_2021}, which leads to blind spots in the
literature. The time-consuming nature of clinical studies causes
delays until information is available to practitioners. Both
limitations can be mitigated by accessing social media
data. \newcite{duh-et-al_2016} find in fact that social media can lead
to earlier detection of adverse drug reactions.

While social media data has come more into focus recently, existing
corpora are limited with respect to the types of entities and
relations they cover. Most commonly, biomedical entity corpora focus
on diseases, symptoms and drugs
\cite[i.a.]{jimeno-yepes_2015,alvaro-et-al_2017}. With regards to
relation detection, work on Twitter is limited to causal relations
\cite{doan-et-al_2019}, or a very small number of relation classes
(i.e.\ \textit{reason-to-use}, \textit{outcome-negative},
\textit{outcome-positive})\ \cite{alvaro-et-al_2017}.  This leaves a
gap for medical information needs. As described
above, content from social media holds this type of
information. Extracting it is required if we want to uncover more
fine-grained aspects of patients' medical journeys complementary to
the knowledge in scientific text.

To facilitate research in this area, we contribute a corpus of medical
tweets annotated with a fine-grained set of medical entities and
relations between them. For the BEAR Corpus of
\textit{\underline{B}iomedical \underline{E}ntities \underline{A}nd
  \underline{R}elations on Twitter}, we annotate 14 entity and 20
relation classes. Entities include environmental factors, diagnostics,
biochemical processes, quality-of-life assessments, pathogens, as well
as more established entity classes such as medical conditions, and
treatments. Relation classes model how entities prevent, influence,
interact with, cause or worsen other entities, or how they relate to
each other as a symptom, side-effect, or diagnosis.

The dataset consists of 2,100 tweets with roughly 6,000 entities and
3,000 relations. To the best of our knowledge the majority of those
classes which are centered around patient journeys have not been
considered before. The dataset is available at
\url{https://www.ims.uni-stuttgart.de/data/bioclaim}.

\begin{table*}
  \centering\small
  \setlength{\tabcolsep}{4pt}
  \renewcommand{\arraystretch}{0.4}

  \renewcommand\tabularxcolumn[1]{m{#1}}
\begin{tabularx}{\textwidth}{lX}

       \toprule
       id& \multicolumn{1}{c}{Tweet}\\
       
		\cmidrule(lr){1-1} \cmidrule(lr){2-2}

		1 & \begin{dependency}[edge style={Gray}, label style={gray!60, fill=gray!60,font=\sf,text=black}]
				\begin{deptext}
					\drug{Prochlorperazine} \&is \drug{compazine}, \&just the generic name. \&\drug{Ativan} \& also causes \&\medC{drowsiness}\\
				\end{deptext}
				\depedge[edge unit distance=0.5ex]{1}{2}{is\_type\_of}
				\depedge[edge unit distance=0.5ex]{6}{4}{side-effect-of}
			\end{dependency} \\
			
		\cmidrule(lr){1-2}
			
		\multirow{2}{*}{} & \begin{dependency}[edge style={Gray}, label style={gray!60, fill=gray!60,font=\sf,text=black}]
				\begin{deptext}[]
					[...] I was on \&\drug{Lyrica}\& to help with the horrific \& \medC{neuropathic pain} \& but cause \& \medC{mind numbing} and \&\medC{bowel problems} \\
				\end{deptext}
				\depedge[edge unit distance=0.6ex]{6}{2}{side-effect-of}
				\depedge[edge unit distance=0.8ex]{7}{2}{side-effect-of}
				\depedge[edge unit distance=0.5ex]{2}{4}{treats}
			\end{dependency} \\
			
			2&\begin{dependency}[edge style={Gray}, label style={gray!60, fill=gray!60,font=\sf,text=black}]
				\begin{deptext}
					so I \& \medC{stopped taking the Lyrica}. \& \qol{I’m in more pain now but feel more like me.} \\
				\end{deptext}
				\depedge[edge unit distance=0.5ex]{2}{3}{cause\_of}
			\end{dependency} \\
			
		\cmidrule(lr){1-2}	
			
		3 &\begin{dependency}[edge style={Gray}, label style={gray!60, fill=gray!60,font=\sf,text=black}]
				\begin{deptext}
				\habit{Meditation}, \& \habit{yoga} \&[-0.0cm] [...] are all effective at relieving \& \medC{stress} \& and helping with \& \medC{\#IBS}.\\ 
				\end{deptext}
				\depedge[edge unit distance=0.5ex]{2}{4}{pos\_influence}
				\depedge[edge unit distance=0.7ex]{2}{6}{pos\_influence}
				\depedge[edge below, edge unit distance=0.5ex]{1}{4}{pos\_influence}
				\depedge[edge below, edge unit distance=0.7ex]{1}{6}{pos\_influence}
				
			\end{dependency} \\
			
		\cmidrule(lr){1-2}	
			
		4 &   \begin{dependency}[edge style={Gray}, label style={gray!60, fill=gray!60,font=\sf,text=black}]
				\begin{deptext}
				\dietary{Alcohol} \& disrupts \& \process{production of adenosine} \& which results in \& \medC{lighter sleep} [...]\\ 
				\end{deptext}
				\depedge[edge unit distance=0.5ex]{1}{3}{neg\_influence}
				\depedge[edge unit distance=0.5ex]{3}{5}{cause\_of}
			\end{dependency} \\
			
		\cmidrule(lr){1-2}	
			
		5 &   \begin{dependency}[edge style={Gray}, label style={gray!60, fill=gray!60,font=\sf,text=black}]
				\begin{deptext}
				\qol{I’m awake just can’t get going.}\& Need cat food seriously only reason go out [...] \& \medC{\#SpoonieLife} \\ 
				\end{deptext}
				\depedge[edge unit distance=0.5ex]{3}{1}{cause\_of}
				\end{dependency} \\
			
		\cmidrule(lr){1-2}	
		
		6 &   \begin{dependency}[edge style={Gray}, label style={gray!60, fill=gray!60,font=\sf,text=black}]
				\begin{deptext}[]
				[...] neighbour has been diagnosed with \& \medC{c19} \& which means admin has to \& \therapy{self isolate} \&and do a \& \diagnostics{test} [...] \\ 
				\end{deptext}
				
				\depedge[edge unit distance=0.5ex]{4}{2}{pos\_influence}				
				\depedge[edge unit distance=0.7ex]{6}{2}{may\_diagnose}
			\end{dependency} \\
			
		\cmidrule(lr){1-2}	
		
		7 &   \begin{dependency}[edge style={Gray}, label style={gray!60, fill=gray!60,font=\sf,text=black}]
				\begin{deptext}
				\other{Support dogs} \& can improve the effectiveness of \&\medC{dementia} \& therapy! Miracle creatures. \\ 
				\end{deptext}
				\depedge[edge unit distance=0.5ex]{1}{3}{pos\_influence}
			\end{dependency} \\

     \bottomrule

\end{tabularx}
\caption{Annotated tweets from the dataset.}
\label{tab:labeled-examples}
\end{table*}

\section{Related Work}
Biomedical natural language processing (BioNLP) is an established
field in computational linguistics, with a rich set of shared tasks
including BioCreative and the competitions organized by the BioNLP
workshop series
\cite{biocreative-proceedings_2021,ben-abacha-et-al_2021}. Research
topics include automatic information extraction from clinical reports,
discharge summaries or life science articles, e.g., in the form of
entity recognition for diseases, proteins, drug and gene names
\cite[i.a.]{Habibi2017,Giorgi2018,Lee2019}. A subsequent task to
entity recognition is relation extraction which covers clinical
relations
\cite{uzuner-et-al_2011,wang-fan_2014,sahu-et-al_2016,Lin2019,akkasi-moensl_2021}
or biomedical relations/interactions (e.g., drug-drug-interactions)
between entities \cite[i.a.]{Lamurias2019,Sousa2021}.

While scientific resources contain high quality information, studies
might not be fully representative regarding population groups or
gender \cite{weber-et-la_2021}, which leads to blind spots in the
literature -- the general population can barely be captured in such
studies. In addition, clinical studies or reports are time-consuming
which inevitably leads to delays, e.g., with regards to indications of
adverse drug events. Both limitations can be mitigated by accessing
social media data. \newcite{duh-et-al_2016} find in fact that social
media can lead to earlier detection of adverse drug reactions.  This
is why biomedical NLP also works with social media texts and online
content \cite[i.a.]{Wegrzyn2011,Yang2016,Sullivan2016}, including
established shared tasks \cite{smm4h-proceedings_2021}. A major focus
has been to inform pharmacovigilance by identifying and extracting
mentions of adverse drug reactions
\cite{nikfarjam-et-al_2015,Cocos2017,magge-et-al_2021}. Additionally,
the community has explored leveraging social media postings to monitor
public health
\cite{Paul2012,Choudhury2013,Sarker2016,stefanidis-et-al_2017}, and
detect personal health mentions
\cite{Yin2015,klein-et-al_2017,Karisani2018}.

A few studies compare biomedical information in scientific documents
with social media: \newcite{Thorne2017} explore how disease names are
referred to across both domains, while \newcite{Seiffe2020} look into
laypersons' medical vocabulary.  A related task is entity
normalization which links a given mention of an entity to the
respective concept in a formalized medical
ontology. \newcite{limsopatham-collier_2016} and later
\newcite{basaldella-et-al_2020} explore this task for medical entities
on social media showcasing the difficulties in mapping laypeople's
health terminology to structured medical knowledge bases.

The ongoing COVID-19 pandemic has sparked bioNLP research to leverage
or contextualize information about the disease and virus from social
media. A number of studies explore detecting COVID-19-related
misinformation and fact-checking
\cite[i.a.]{hossain-et-al_2020,chen-hasan_2021,mattern-et-al_2021,saakyan-et-al_2021}. Others
have looked into monitoring information surrounding the virus using
social media \cite{cornelius-et-al_2020,hu-et-al_2020}.

\subsection{BioNER on Social Media}
Early contributions on biomedical information extraction from Twitter
aimed at the extraction of adverse drug reactions from social media --
a fundamentally different use case than scientific text analytics. The
goal is to provide access to information even before it becomes
available to doctors or researchers. This work includes corpus creation
efforts on dedicated platforms like
AskAPatient\footnote{\url{https://www.askapatient.com/}}
\cite{karimiet-al_2015} and Twitter
\cite{nikfarjam-et-al_2015,magge-et-al_2021}

With a similar motivation, \newcite{jimeno-yepes_2015} created
Micromed, a Twitter corpus annotated with disease names, drug names,
and symptom mentions. Further, TwiMed \cite{alvaro-et-al_2017} is a
dataset which combines social media and scientific text with
annotations of diseases, symptoms and drug names to study drug reports
across both sources. Annotated with the same entity classes, the
MedRed dataset consists of Reddit posts \cite{scepanovic-et-at_2020}
labeled via crowdsourcing.

In addition to identifying entities, there has also been some work on
linking them to existing databases. To facilitate this task for social
media, \newcite{limsopatham-collier_2016} contribute a Twitter corpus
in which entities are linked to the SIDER~4\cite{sider_2016} database
of drug profiles. \newcite{basaldella-et-al_2020} subsequently
introduce COMETA, a Reddit corpus in which entities are linked to
SNOWMED-CT\footnote{\url{https://www.snomed.org/}}. With regards to
the groups of entities considered (phenotype, disease, anatomy,
molecule (incl. drugs, toxins, nutrients etc.), gene/DNA/RNA, device,
procedure) this is similar to our contribution.

Existing resources do not cover enough entities to extract patient
narratives from social media. They do not allow us yet to access the
fine-grained information that social media content holds, and that
would allow us to fill the information gap in scientific text.

\subsection{Detection of Medical Relations on Social Media}
Relation extraction contextualizes entities with each other.
Medical relation extraction resources for social media are rare. Existing studies have focused on causal relations
\cite{doan-et-al_2019}, or a small number of relation classes
(i.e., \textit{reason-to-use}, \textit{outcome-negative},
\textit{outcome-positive})\ \cite{alvaro-et-al_2017}.

With regards to scientific text, and specifically clinical relation
extraction, closest to our annotation scheme are approaches by
\newcite{uzuner-et-al_2011} and \newcite{wang-fan_2014}. Classes for both their work describe
relations between treatments and medical conditions, relations between
two treatments, medical conditions, or diagnoses (e.g.,
\textit{treatment caused medical problem}, \textit{treatment improved
  or cure medical problem}, \textit{test reveal medical problem} in 
\newcite{uzuner-et-al_2011}, or \textit{treats},
\textit{prevents}, \textit{has symptom}, \textit{contraindicates} in
\newcite{wang-fan_2014}).  However, both work with clinical and
scientific texts. Medical relation extraction on social media is
understudied and missing resources that facilitate extracting
patients' experiences and opinions towards entities of their medical
history which would allow us to recover their medical narratives.

\section{Corpus Creation}

\subsection{Data Collection}
\label{sec:data-collection}
We collect English tweets between January 01 and November 02, 2021 using the
official keyword-based Twitter API.

\subsubsection{Corpus Subselection}
The list of keywords to retrieve the data stems from three different
sources. Refer to Table \ref{tab:search-term-examples} for examples
for each source.
\begin{compactenum}
\item DrugBank: DrugBank is a database for drugs which provides
  molecular information about drugs, their mechanisms, interactions
  and targets \cite{drugbank_2018}. We use generic and brand/product
  names which allows us to collect tweets discussing treatments, or
  descriptions of off-label drug use.
\item MeSH: Medical Subject Headings is a controlled vocabulary
  thesaurus used for indexing articles in PubMed\footnote{\url{https://pubmed.ncbi.nlm.nih.gov/}}. We use terms from
  the subcategories disease and therapeutics to collect
  tweets that address specific diseases and therapeutic measures. We
  use all terms that appear with a frequency $>=$ 1000 in
  PubMed articles
  hypothesizing that the distribution of those terms mirrors the usage
  on Twitter.
\item Manual: MeSH and DrugBank mostly contain scientific terms (see
  Table \ref{tab:search-term-examples}), so we also query with a
  manually compiled list of medical terms. Partly, those relate to 10
  medical conditions\footnote{COVID-19, Alzheimer's disease,
    borderline personality disorder, cancer, depression, irritable
    bowel syndrome, measles, multiple sclerosis, post-traumatic stress
    disorder, stroke.}. This is to collect tweets that either use
  Twitter specific hashtags, abbreviations, or community-based terms
  related to a condition, or mention terms generally related to the
  medical domain.
  
\end{compactenum}
All terms combined result in a list of 22,874 keywords. From this
list, 10,599 terms return results from Twitter during a test crawl. We
remove unproductive terms and use a final list of 7,358 keywords from
DrugBank, 3,120 from MeSH, and 121 from the manually compiled
list.\footnote{Lists we used to collect and filter the data are available in the
  suppl.\ material together with the corpus.} We acknowledge that by using this approach, we can not sample tweets with incorrectly spelled mentions of drug or disease names.

We only keep non-duplicate tweets (based on the tweet ID) which do not
contain a URL due to their increased probability of containing
advertisements. Further, we only keep tweets which contain a
relational term. Examples include words like \textit{treats},
\textit{prescribed}, or \textit{diagnosed} (and variations
thereof). From the resulting collection of tweets, we draw a sample
balanced across the three keyword sources. We subsequently annotate
700 tweets per data source (350 per MeSH subcategory) which amounts in
a total of 2,100 tweets.

\subsection{Annotation}
We label entity and relation classes that allow us to include
individual aspects within people's disease-treatment cycles. Classes
cover information concerning developing symptoms, being diagnosed and
offered a treatment, being cured or learning to live with a medical
condition. They allow us to model statements about how to
self-diagnose, treat a particular condition by themselves, or capture
how people perceive risk factors.  For both annotation tasks, we
therefore follow the central paradigm which tells annotators to label
entities and relations the way a tweet's author intends or understands
them. A mention like \textit{UV radiation} could either be intended as
an environmental factor (High UV radiation causes skin cancer.), or a
treatment (UV radiation will help with my low vitamin D levels).

\begin{figure}
  \centering
  \includegraphics[scale=0.8,page=3]{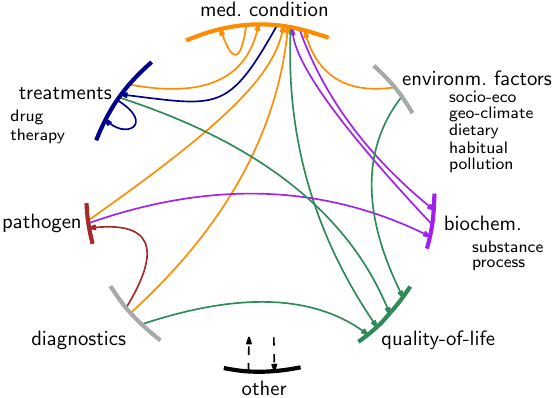}
  \caption{Visualization of entity classes and the relations between
    them.}
  \label{fig:relationvisualization}
\end{figure}

\subsubsection{Entity classes}
\label{sec:intro-ent-classes}

We label seven groups of entities. Each group contains a respective
label or subset of labels which the annotators use to label the
text. We visualize the entities in
Figure~\ref{fig:relationvisualization} and depict which entity-pairs
can be related.  Each entity group will be briefly described in the
following section. Table \ref{tab:labeled-examples} additionally
provides fully annotated examples from the dataset, to which we will
refer to in the following descriptions.

\paragraph{Medical Conditions.}
All mentions of diseases, symptoms, side
effects, and medical events or descriptions
thereof. See \medC{\#IBS} in Ex.~3 or \medC{drowsiness} in Ex.~1. Phrases like
\medC{stopped taking the Lyrica} in Ex.~2 are considered
relevant medical events, and labeled as \textit{medC}, too.

\paragraph{Treatments.}
Mentions of any kind of treatment. That
includes drug names, generic and brand names (see
\drug{Prochlorperazine} and \drug{compazine} in Ex.~1) and all types of therapy or
prevention methods (see \therapy{self isolate} in Ex.~6).

\paragraph{Environmental Factors.}
Entities that influence, cause or
contribute to a medical condition. We annotate
socio-economic (age, gender, ethnicity, social background etc.),
geographic/climatic (geography, climate, weather etc.), dietary,
habitual (exercise, stress etc.) or pollution-related (air/water pollution, UV or nuclear radiation)
factors. See \habit{yoga} in Ex.~2 or \dietary{Alcohol} in Ex.~4.

\paragraph{Pathogens.}
Pathogens are organisms that cause diseases. This includes mentions of
bacteria, fungi, parasites, or viruses, e.g.,
\pathogen{coronavirus}.

\paragraph{Biochemical Entities.}
Biochemical substances such as proteins or hormones (e.g.,
\substance{Lactose}). The class includes biochemical processes such as
biological, pathogenic or chemical mechanisms (see \process{production
  of adenosine} in Ex.~4).

\paragraph{Diagnostics.}
Mentions of tests or other diagnostic instruments that are used to diagnose or test for a medical condition. Refer to \diagnostics{test} in Ex.~6.

\paragraph{Quality of Life Assessments.}
Descriptions of patients' quality of life, i.e. mentions of how a disease or its management impacts a patient's well-being. See \qol{I'm awake just can't get going} in Ex.~5.

\paragraph{Other.} Relevant entities that can not be covered by any of the other classes. See \other{Support dogs} in Ex.~7.

\subsubsection{Relation Classes}
\label{sec:intro-rel-classes}

Each relation is directed and connects two entities (see Figure
\ref{fig:relationvisualization} for a depiction of which entities can be related and Table~\ref{tab:labeled-examples} for examples). We annotate the following entity
pairs with relations. ($\pm$ indicates that a relation has a positive and negative variant, e.g., (does not) treat.)

\begin{compactdesc}
\item[treat$\,\to\,$medC] $\pm$treats, worsens, $\pm$prevents, $\pm$causes, contraindicates, prescribed, $\pm$influences
\item[medC$\,\to\,$treat] side effect of
\item[env/pathogen/biochem$\,\to\,$medC] $\pm$causes, $\pm$influences, $\pm$prevents 
\item[medC$\,\to\,$medC/biochem] has symptom,  $\pm$causes, is similar to
\item[treat$\,\to\,$treat]  $\pm$interaction, is similar to
\item[diag$\,\to\,$medC/pathogen]  $\pm$diagnoses
\item[pathogen$\,\to\,$biochem]  $\pm$causes
\item[medC/treat/env/diag$\,\to\,$qol]  $\pm$causes, $\pm$influences
\item[general] type of, other 
\end{compactdesc}

\subsubsection{Evaluation metrics}
\label{corpus-eval-metrics}

We measure the agreement between annotations by calculating the
inter-annotator \F. Specifically, we treat one annotator's labels as
the gold annotations and consider the other annotator's labels as
predictions \cite{hripcsak-rothschild_2005}.

We report the agreement for varying levels of
strictness. We consider
entity span (S) and type (T) as follows:
\begin{compactitem}
\item [S1T1] The two spans and types of the entities are entirely identical.
\item [S0T1] The two spans overlap by min. one token, entity type is identical.
\item [S0T0] The two spans overlap by min. one token, entity type
  is ignored in the comparison.
\end{compactitem}

When evaluating the annotated relation (R) between two entities, we consider two modes:
\begin{compactitem}
\item [R1] Relation type and direction are identical.
\item [R0] Relation type and direction are ignored in the comparison.
\end{compactitem}

On the entity level, comparing S1T1 to S0T1 shows to which extend
the span of an entity influences the annotation task. Comparing
S0T1 to S0T0 indicates the impact of assigning a label on the
difficulty of the task.

Analyzing the relation annotation follows the same objectives with
respect to the entities, but adds the impact of the relation
assignment. R1S1T1 is the strictest evaluation mode. The comparison to both R1S0T1 and R1S0T0 helps in
understanding how the entity annotation influences the relation
annotation task. R0S0T0 captures the most general level of agreement
indicating how well the annotators can identify the fact that any two
entities are somehow related. Comparing this to R1S0T0,
we can conclude how difficult it is to identify relation types.

\subsubsection{Guideline development \& annotator training}
We work with two in-house annotators (A1, A2) to label the tweets with entities
and relations. Both annotators are female, ages 20 to 25, and 25 to 30, respectively. Their
backgrounds are in linguistics and computational linguistics. They have no
medical training. We iteratively train the annotators over the course of three
months. In each training iteration, all
annotators label a small set of instances independently following our
annotation guidelines. Subsequently we discuss each set within the
group. In addition, we calculate the inter-annotator \F for each round
of training annotations (refer to Section \ref{corpus-eval-metrics}
for an explanation of the eval. metrics used), and adapt the
guidelines with findings from the discussions and analysis to clarify
the annotation tasks further. The training instances are not part of
the final corpus. The final version of the guideline document is
available in the supplementary material.
 
Table \ref{tab:anno-training} shows the development of the inter-annotator \F over the training iterations. For each round we report the macro \F score across all entity/relation classes in the different evaluation settings.
We find that the agreement increases for the entities and the relation annotation over time.
The agreement increases as we allow for less precise matches to be counted as true positive instances.
By the end of the training period, annotators agreed with .53\F on exact entity types and boundaries (S1T1). Comparing the impact of each subtask in the last round, we observe that agreeing on the entity type is more challenging than identifying the entity span (decrease of .25\F between S0T0 and S0T1 vs. .02\F decrease between S0T1 and S1T1).
Evaluating the relation type strictly (R1S0T0 vs. R0S0T0), the agreement drops by .07\F which indicates that the relation type is fairly ambiguous, and therefore hard to agree upon. 
The strictest evaluation measures (S1T1, R1S1T1) show that the task remains challenging even after substantial annotator training which we attribute to the diverse nature of text in tweets.
Presumably, this is also why the agreement fluctuates over training rounds.

\begin{table}
\centering\small
\begin{tabular}{rrrrrrrr}

      \toprule
      
      & \multicolumn{7}{c}{Evaluation mode} \\
      \cmidrule(lr){2-8}
      Round
      & \rotatebox{90}{S1T1}  
      & \rotatebox{90}{S0T1} 
      & \rotatebox{90}{S0T0} 
      & \rotatebox{90}{R1S1T1} 
      & \rotatebox{90}{R1S0T1} 
      & \rotatebox{90}{R1S0T0}
      & \rotatebox{90}{R0S0T0}\\
		
		\cmidrule(r){1-1} \cmidrule(lr){2-2} \cmidrule(lr){3-3}  \cmidrule(lr){4-4} \cmidrule(lr){5-5} \cmidrule(lr){6-6} \cmidrule(lr){7-7} \cmidrule(lr){8-8}              
        
1 & .44 & .66 & .73 & .18 & .4 & .4 & .4\\
2 & .23 & .34 & .75 & .18 & .25 & .35 & .35\\
3 & .37 & .45 & .79 & .05 & .23 & .27 & .59\\
4 & .64 & .76 & .95 & .44 & .48 & .51 & .68\\
5 & .39 & .46 & .76 & .07 & .1 & .29 & .43\\
6 & .44 & .49 & .89 & .12 & .3 & .44 & .58\\
7 & .62 & .68 & .96 & .4 & .4 & .44 & .47\\
8 & .42 & .61 & .77 & .15 & .39 & .41 & .56\\
9 & .69 & .77 & .82 & .31 & .33 & .35 & .54\\
10 & .53 & .55 & .8 & .28 & .4 & .59 & .66\\

            \bottomrule

\end{tabular}
\caption{Macro inter-annotator \F across all entity and relation classes throughout the training rounds. S1T1 through R0S0T0 indicate the evaluation mode.}
 \label{tab:anno-training}
\end{table}

\subsection{Aggregation}
We provide an adjudicated version of the dataset which combines both
annotators' results.  In case of disagreements of entity
spans between the annotators, we choose the longest overlapping
sequence between two instances. We further prefer more frequent entity
and relation classes over less frequent ones, and choose more general
concepts over more specific ones. Generally, our aggregation strategy
is motivated by a high recall approach to ensure that we lose as
little of the nuances from the individual annotations as possible.  We
aggregate in two steps and first align the entity annotations,
followed by aggregating the relations.  Please refer to
Section~\ref{appendix:aggregation} for more details.

\section{Analysis}
\subsection{Agreement Between Annotators}
The annotators labeled the final corpus over the course of four
months.  
Since both sets of annotations provide unique perspectives on the
data, we release the individual
annotations along with an aggregated version.
We evaluate the annotations using the inter-annotator
\F-scores as described in Section \ref{corpus-eval-metrics} and
provide scores for the full dataset as well as individual scores for
each sampling method in the following.
\begin{figure}
\centering
\includegraphics[scale=0.5]{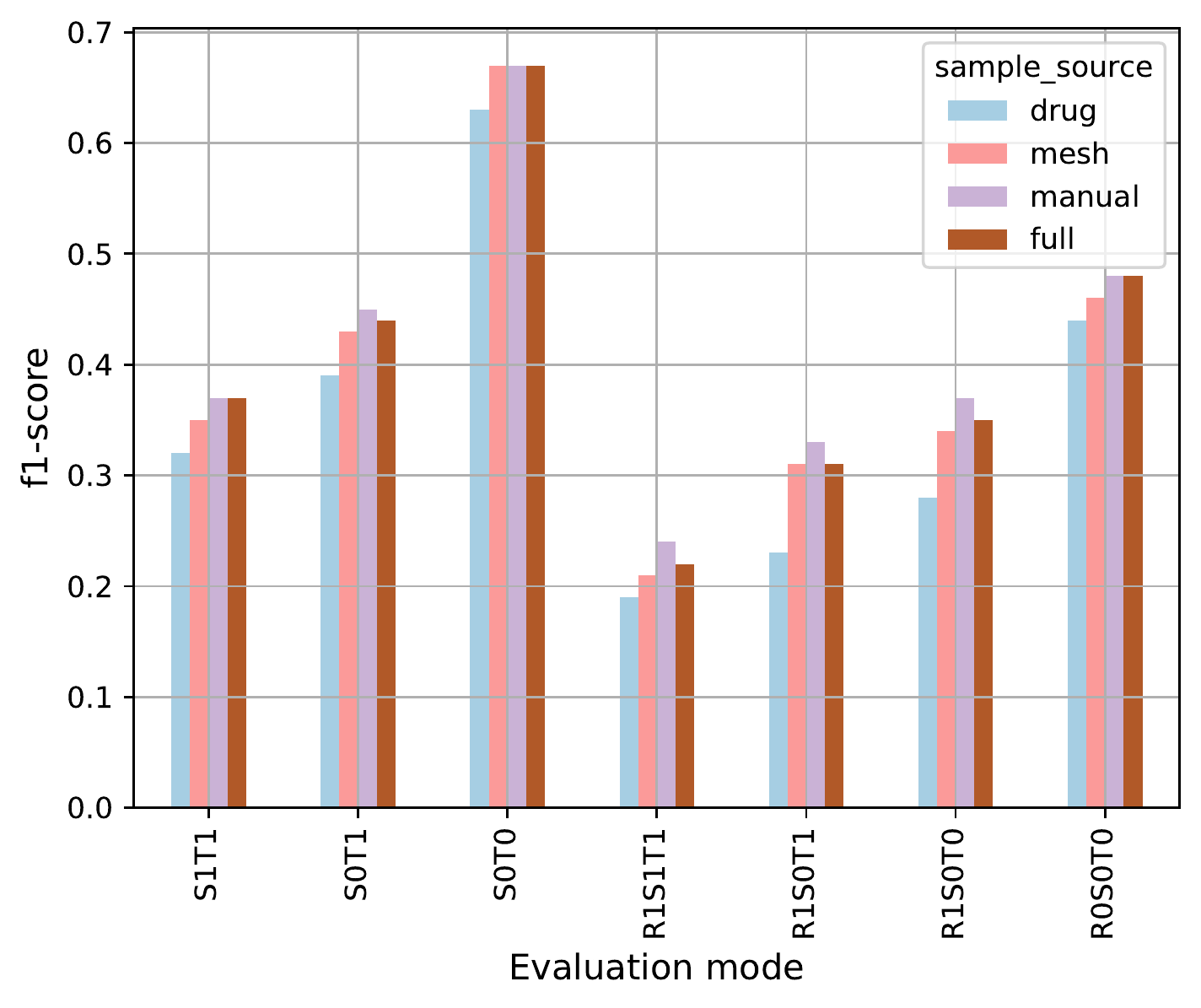} 
\caption{Inter-annotator macro \F-scores for each subsample of the
  corpus (DrugBank (drug), MeSH (mesh), manually researched keywords (manual)), and the full dataset
  (full) across evaluation modes.}
\label{fig:agreement-scores-per-subsample-and-mode}
\end{figure}
Figure \ref{fig:agreement-scores-per-subsample-and-mode} shows the
inter-annotator \F-scores for each subsample of the corpus
evaluated with descending strictness.
For the final corpus we find that
annotators are fairly synchronized in identifying 
entities in tweets (.67\F S0T0). Agreeing on the entity type is more challenging than identifying
the same entity span (.07\F decrease between S0T1 and S1T1 vs. .23\F
decrease between S0T0 and S0T1). This is also the case for the relation agreement.
Labeling the relation type is by far the most difficult task. When we compare
the agreement levels in R0S0T0 with R1S0T0, we report a
difference of .13\F which showcases how ambiguous the relations are.

We observe a slight decrease of the agreement compared to the last
training round.  We attribute this to the fact that annotators are
continued to be faced with novel variations of entities and relations
because of Twitter's diverse nature.

\paragraph{Agreement across sources.} Across all evaluation modes, tweets from the subsample Manual show the strongest agreement, followed by subsamples MeSH, and DrugBank.
The results indicate that tweets from the Manual category are
easier to annotate than the other documents, presumably because they mostly use laypeople's vocabulary. Due to the nature of the DrugBank database, tweets from this set might be more scientific, making them more difficult to
annotate. 

\paragraph{Agreement across entities.}
Table \ref{tab:entity-stats} reports the inter-annotator \F-score
(iaa) for each entity class (eval. mode: S1T1).  A1 and A2 agree most strongly on
instances of \textit{medC} and \textit{treat\_drug} (.73 and .74 \F,
respectively). We observe the lowest agreement for mentions of
\textit{biochem\_process} (.05 \F).

We observe that the agreement for highly frequent classes is stronger
than the agreement in less frequent ones. Presumably, this is because
these classes are also the most concrete, and therefore easier to
detect. Less frequent classes (e.g., \textit{env} or
\textit{qol}) could be considered more abstract or vague. At the same
time, we presume that seeing a certain type of entity more often acts
like a training effect for the annotators.

\paragraph{Agreement across relations.}
Table \ref{tab:relation-stats} reports the inter-annotator \F-score for each relation class (evaluation mode: R1S0T0\footnote{We choose R1S0T0 to focus specifically on the relation while allowing a imprecise agreement on the entities.}). Across all classes, we report a macro \F-score of .35. \textit{has\_symptom}  and \textit{does\_not\_prevent} are the classes with highest agreement (.59 \F respectively), followed by \textit{treats} (.58 \F), \textit{may\_diagnose} and \textit{prevents} (.56 \F, respectively). We observe no agreement for \textit{is\_contraindicated}, \textit{may\_not\_diagnose}, and \textit{pos/neg\_interaction}. 

\subsection{Corpus Statistics}
The final corpus contains 2,100 tweets with labels for medical
entities and the relations connecting them.  Table
\ref{tab:general-class-distribution} lists the number of documents
with and without entities and relations. The
majority of documents in the dataset contain entities. 86.2 \% of all documents in the dataset are labeled with at least one
entity. Slightly
more than half of all documents containing entities also express a
relevant relation (56.5 \%).

The corpus consists of 93,258 words (17,559 words are
unique). The longest tweet consists of 114 words, the two shortest
tweets are made up of 4 words each (see Table
\ref{tab:shortest-longest-tweet}). A tweet from our corpus has an
average length of 44.41 words. There is no substantial difference
between tweets from different sampling sources.

The following sections describe our dataset in more detail. We present
corpus statistics regarding the entity and relation class
distribution. Note that we describe the aggregated version of the dataset.

\begin{table}
  \centering\small
  \setlength{\tabcolsep}{5pt}
\begin{tabularx}{\columnwidth}{lrrrrr}
  \toprule
  & \multicolumn{4}{c}{Number of documents} \\
  \cmidrule(lr){2-6}
  & no ent & with ent & no rel & with rel \\
  \cmidrule(lr){2-2} \cmidrule(lr){3-3} \cmidrule(lr){4-4} \cmidrule(lr){5-5}
  A1 & 330 (15.7) & 1770 (84.3) & 835 (47.2) & 935 (52.8) \\
  A2 & 378 (18.0) & 1722 (82.0) & 833 (48.4) & 889 (51.6) \\
  \cmidrule(lr){2-5}
  agg &289 (13.8) & 1811 (86.2) & 788 (43.5) & 1023 (56.5) \\
 
  \bottomrule
\end{tabularx}
\caption{Number of documents with and without entities (ent) and
  relations (rel) for both annotators (A1, A2) and the aggregated dataset (agg). Values in parenthesis report the respective percentages. For relations this is w.r.t. all instances which contain entities.}
 \label{tab:general-class-distribution}
\end{table}

\subsubsection{Entities}
\label{corpus-stats-entities}

Table \ref{tab:entity-stats} shows the number of instances
per entity class. We include the statistics for both annotators (A1,
A2) and for the adjudicated dataset. Additionally, we report the statistics for the whole corpus (full), and divided by the method the documents were sampled with (DrugBank, MeSH terms, Manual) .

The dataset contains 6,324 entities.
The biggest entity class is medical conditions (3,553 instances),
followed by mentions of \textit{treat\_drug} (1,240).  The remaining
entity classes are substantially less frequent. \textit{env\_pollution} has the smallest number of instances (5).  Annotators label approx. 3.01 entities per document.

\paragraph{Entities across sources.}
Mentions of medical conditions are more frequent in tweets from the subsamples MeSH and Manual (1,458 and 1,367, respectively) than they are the DrugBank sample (728). Tweets from set DrugBank exhibit the majority of mentions of \textit{treat\_drug} as well as \textit{biochem\_substance} entities (1,035 and 163, respectively). Notably, mentions of the second treatment-related entity class, \textit{treat\_therapy}, are more frequent in tweets from the MeSH and Manual sample.

These results confirm that tweets in the DrugBank sample more frequently discuss treatments, and therefore exhibit a high number of drug and biochemical entities. 
\textit{treat\_therapy} captures more general treatment descriptions than specific mentions of drugs. Regarding the subsample Manual, we presume that the high frequency of therapy mentions indicates that laypeople speak in more general terms about treatments.

\begin{table*}
\centering\small
\setlength{\tabcolsep}{6.0pt}
\begin{tabular}{lrrrrrrrrrrrrrrrr}

       \toprule
       &\multicolumn{15}{c}{Entities}\\
       \cmidrule(lr){2-17} 
   
       &\rotatebox{90}{medC} 
       & \rotatebox{90}{treat\_drug} 
       & \rotatebox{90}{treat\_therapy} 
       & \rotatebox{90}{env\_geo-cli} 
       & \rotatebox{90}{env\_diet} 
       & \rotatebox{90}{env\_habit} 
       & \rotatebox{90}{env\_pollution} 
       & \rotatebox{90}{env\_socio-econ} 
       & \rotatebox{90}{pathogen} 
       & \rotatebox{90}{qol} 
       & \rotatebox{90}{diag} 
       & \rotatebox{90}{biochem\_process}
       & \rotatebox{90}{biochem\_subst} 
       & \rotatebox{90}{other} 
       & total
       & \rotatebox{90}{av. \#ents/doc}\\
       
		\cmidrule(lr){2-15} \cmidrule(lr){16-16} \cmidrule(lr){17-17}

		\multirow{3}{*}{\rotatebox{90}{~DB~}} 
			& 674 & 1053 & 99 & 1 & 47 & 4 & 1 & 3 & 14 & 23 & 23 & 24 & 139 & 48 & 2153 & 3.08 \\
			& 620 & 967 & 84 & 0 & 43 & 2 & 1 & 6 & 13 & 22 & 8 & 19 & 189 & 39 & 2013 & 2.88\\ 
			\cmidrule(lr){2-15} \cmidrule(lr){16-16} \cmidrule(lr){17-17}
			& 728 & 1035 & 114 & 0 & 34 & 2 & 1 & 5 & 7 & 28 & 13 & 6 & 163 & 41 & 2177 & 3.11\\
			
		\cmidrule(r){1-17}		
		
		\multirow{3}{*}{\rotatebox{90}{~MeSH~}} 
			&  1361 & 151 & 366 & 2 & 42 & 23 & 2 & 8 & 63 & 11 & 40 & 12 & 43 & 45 & 2169 & 3.1\\ 
			&  1288 & 108 & 324 & 3 & 18 & 23 & 1 & 20 & 45 & 15 & 25 & 8 & 39 & 58 & 1975 & 2.82\\ 
			\cmidrule(lr){2-15} \cmidrule(lr){16-16} \cmidrule(lr){17-17}
			& 1458 & 128 & 347 & 3 & 36 & 25 & 2 & 19 & 35 & 12 & 27 & 3 & 38 & 43 & 2176 & 3.11\\ 
			
		\cmidrule(r){1-17}			
		
		\multirow{3}{*}{\rotatebox{90}{Manual}} 
			& 1329 & 89 & 254 & 0 & 29 & 30 & 3 & 10 & 36 & 38 & 33 & 4 & 18 & 49 & 1922 & 2.75\\
			& 1276 & 60 & 226 & 3 & 29 & 27 & 0 & 40 & 41 & 64 & 21 & 14 & 31 & 45 & 1877 & 2.68\\ 
			\cmidrule(lr){2-15} \cmidrule(lr){16-16} \cmidrule(lr){17-17}
			& 1367 & 77 & 234 & 3 & 34 & 21 & 2 & 28 & 30 & 67 & 28 & 7 & 22 & 51 & 1971 & 2.82\\
			
		\cmidrule(r){1-17}		
		
		\multirow{3}{*}{\rotatebox{90}{~full~}} 
			& 3364 & 1293 & 719 & 3 & 118 & 57 & 6 & 21 & 113 & 72 & 96 & 40 & 200 & 142 & 6244 & 2.97\\   
			& 3184 & 1135 & 634 & 6 & 90 & 52 & 2 & 66 & 99 & 101 & 54 & 41 & 259 & 142 & 5865 & 2.79 \\ 
			\cmidrule(lr){2-15} \cmidrule(lr){16-16} \cmidrule(lr){17-17}
			& 3553 & 1240 & 695 & 6 & 104 & 48 & 5 & 52 & 72 & 107 & 68 & 16 & 223 & 135 & 6324 & 3.01 \\  
			
		\cmidrule(r){1-17}

		iaa
			& .73 & .74 & .66 & .22 & .36 & .39 & .25 & .18 & .43 & .15 & .44 & .05 & .42 & .12 & \multicolumn{2}{c}{.37} \\
       
       \bottomrule

\end{tabular}
\caption{Number of annotated entities and inter-annotator \F (iaa) per entity class. We report the statistics across the whole corpus (full) as well as divided by the method the documents were sampled with (DB $=$ DrugBank, MeSH $=$ Medical subject headings, Manual $=$ manually compiled medical keywords). Within each sampling method we
  report the statistics for annotator 1 and 2 and for the adjudicated dataset. Reported agreement scores (iaa, eval. mode S1T1) for all instances across the full corpus.}
\label{tab:entity-stats}
\end{table*}

\begin{table*}
  \centering\small
  \setlength{\tabcolsep}{4.0pt}
  \renewcommand{\arraystretch}{0.999}

\begin{tabular}{lrrrrrrrrrrrrrrrrrrrrrrr}

       \toprule
       &\multicolumn{17}{c}{Relations}\\
       
       \cmidrule(lr){2-23}	
   
        &\rotatebox{90}{cause\_of}
        &\rotatebox{90}{does\_not\_prevent}
		&\rotatebox{90}{does\_not\_treat}
		&\rotatebox{90}{has\_symptom}
		&\rotatebox{90}{is\_contraindicated}
		&\rotatebox{90}{is\_similar\_to}			
		&\rotatebox{90}{is\_type\_of}
		&\rotatebox{90}{may\_diagnose}
		&\rotatebox{90}{may\_not\_diagnose}
		&\rotatebox{90}{neg\_influence\_on}
		&\rotatebox{90}{neg\_interaction}
		&\rotatebox{90}{not\_cause\_of}
		&\rotatebox{90}{pos\_influence\_on}
		&\rotatebox{90}{pos\_interaction}
		&\rotatebox{90}{prescribed\_for}
		&\rotatebox{90}{prevents}
		&\rotatebox{90}{side\_effect\_of}
		&\rotatebox{90}{treats}
		&\rotatebox{90}{worsens}
		&\rotatebox{90}{other} 
		& total 
		&\rotatebox{90}{~av. \#rels/doc~} \\

		\cmidrule(lr){2-21} \cmidrule(lr){22-22} \cmidrule(lr){23-23}	
		  
		\multirow{3}{*}{\rotatebox{90}{~DB~}} 
			& 215 & 1 & 13 & 3 & 3 & 8 & 138 & 8 & 0 & 49 & 12 & 7 & 67 & 29 & 9 & 27 & 35 & 197 & 1 & 30 & 852 & 1.22\\
			& 157 & 1 & 16 & 12 & 7 & 6 & 92 & 4 & 0 & 31 & 1 & 13 & 75 & 0 & 7 & 29 & 41 & 212 & 1 & 13 & 718 & 1.03\\ 
			\cmidrule(lr){2-21} \cmidrule(lr){22-22} \cmidrule(lr){23-23}	
			& 245 & 1 & 15 & 11 & 4 & 1 & 163 & 7 & 0 & 50 & 11 & 12 & 88 & 28 & 8 & 31 & 52 & 277 & 0 & 39 & 1043 & 1.49\\
			
       \cmidrule(r){1-23}
		 
		 \multirow{3}{*}{\rotatebox{90}{~MeSH~}} 
			& 324 & 0 & 10 & 81 & 5 & 3 & 81 & 14 & 1 & 45 & 2 & 10 & 69 & 3 & 1 & 30 & 23 & 132 & 1 & 10 & 845 & 1.21\\ 
			& 242 & 0 & 12 & 90 & 1 & 6 & 51 & 9 & 2 & 74 & 0 & 20 & 61 & 0 & 0 & 44 & 38 & 121 & 1 & 6 & 778 & 1.11\\ 
			\cmidrule(lr){2-21} \cmidrule(lr){22-22} \cmidrule(lr){23-23}	
			& 407 & 0 & 11 & 108 & 5 & 5 & 102 & 16 & 3 & 48 & 1 & 18 & 92 & 3 & 0 & 40 & 39 & 166 & 1 & 16 & 1081
& 1.54\\ 
			
       \cmidrule(r){1-23}			 
		 
		 \multirow{3}{*}{\rotatebox{90}{Manual}} 
			& 292 & 7 & 7 & 36 & 0 & 13 & 69 & 20 & 1 & 36 & 2 & 7 & 62 & 1 & 0 & 28 & 3 & 48 & 0 & 35 & 667 & 0.95\\
			& 257 & 8 & 8 & 62 & 0 & 8 & 24 & 17 & 0 & 91 & 0 & 17 & 80 & 0 & 0 & 31 & 9 & 40 & 0 & 8 & 660 & 0.94\\ 
			\cmidrule(lr){2-21} \cmidrule(lr){22-22} \cmidrule(lr){23-23}	
			&331 & 9 & 10 & 57 & 0 & 13 & 71 & 23 & 1 & 78 & 0 & 15 & 83 & 1 & 0 & 41 & 10 & 57 & 0 & 35 & 835
& 1.19\\

       \cmidrule(r){1-23}			
		
		\multirow{3}{*}{\rotatebox{90}{~full~}} 
			& 831 & 8 & 30 & 120 & 8 & 24 & 288 & 42 & 2 & 130 & 16 & 24 & 198 & 33 & 10 & 85 & 61 & 377 & 2 & 75 & 2364 & 1.13\\   
			& 656 & 9 & 36 & 164 & 8 & 20 & 167 & 30 & 2 & 196 & 1 & 50 & 216 & 0 & 7 & 104 & 88 & 373 & 2 & 27 & 2156 & 1.03\\
 \cmidrule(lr){2-21} \cmidrule(lr){22-22} \cmidrule(lr){23-23}	
			& 983 & 10 & 36 & 176 & 9 & 19 & 336 & 46 & 4 & 176 & 12 & 45 & 263 & 32 & 8 & 112 & 101 & 500 & 1 & 90 & 2959 & 1.41\\ 
			
       \cmidrule(r){1-23}
		iaa
			& .48 & .59 & .52 & .59 & .0 & .55 & .45 & .56 & .0 & .25 & .0 & .24 & .38 & .0 & .12 & .56 & .54 & .58 & .5 & .08 & \multicolumn{2}{c}{.35}\\
			
       \bottomrule

\end{tabular}
\caption{Number of annotated relations and inter-annotator \F (iaa)
  per class. We report the statistics across the whole corpus (full)
  as well as divided by the method the documents were sampled with (DB
  $=$ DrugBank, MeSH $=$ Medical subject headings, Manual $=$ manually
  researched medical keywords). Within each sampling method we
  report the statistics for annotator 1 and 2 and for the aggregated dataset. Reported agreement scores (iaa, eval. mode R1S0T0) for all instances across the full corpus.}
\label{tab:relation-stats}
\end{table*}

\subsubsection{Relations}
\label{corpus-stats-relations}
Table \ref{tab:relation-stats} reports the number of annotated relations for each class. We calculate the statistics for both annotators (A1, A2) and for the adjudicated data. We report the numbers of relations for the full corpus as well as for each of the three subsamples (DrugBank, MeSH, Manual).

In total, the corpus contains 2,959 relations.
The \textit{cause\_of} relation is the most frequent (983), followed
by \textit{treats} (500), \textit{is\_type\_of} (336), and
\textit{pos\_influence} (263).
\textit{worsens} is the class with the lowest frequency (1 instance).
For relations which can be either positive or negative, the negative relations are always less frequent.
On average, a document in our dataset contains 1.41 relations.

\paragraph{Relations across sources.}
While documents from the subsample DrugBank and MeSH show relatively equal numbers of total relations (averages of 1,043 and 1,081, respectively), the Manual subsample has the least amount of relations (av. of 835).
\textit{cause\_of} relations are most frequent in the subsamples MeSH (407) and Manual (331). In the DrugBank set, \textit{treats} is the most prevalent relation class (277). Notably, for set Manual, we find that \textit{cause\_of} is by far more frequent than any other relation. All other classes count (mostly substantially) less than 100 instances each.

\section{Conclusion and Future Work}

We introduce and describe BEAR, a corpus of 2,100 medical tweets
annotated with a detailed set of biomedical entities, and the
relations connecting them. Both the entity and relation classes are
motivated by the need to capture fine-grained aspects of patients'
medical journeys. In our annotation study, we show that tweets hold
this type of information, and that non-expert annotators can detect
this reasonably well.

With this dataset, we lay the groundwork to develop entity and
relation extraction systems that give medical professionals access to
patient narratives which are not covered in scientific texts. This includes quality-of-life assessments,
perception of risk factors, unconventional treatments, or
self-diagnoses that people might feel uncomfortable or irrelevant to
share with their doctors.
Such systems could help answer detailed questions like ''How does chemotherapy affect the social life of breast cancer patients?" or ''Which habits serve as coping mechanisms for people suffering from depression?".

\subsection*{Acknowledgments}
This research has been conducted as part of the FIBISS project
which is funded by the German Research Council (DFG, project number: KL 2869/5-1). We thank our annotators for their hard work and tireless attention to detail.

\clearpage

\section{Bibliographical References}\label{reference}
\bibliographystyle{lrec2022-bib}
\bibliography{literature}


\appendix
\section*{Appendix}

\subsection*{Example Terms from the Sampling Methods}
Table \ref{tab:search-term-examples} shows terms from each sampling method.
\begin{table}[h!]
\centering\small
\begin{tabularx}{\columnwidth}{lX}
      \toprule
       Source & Example terms \\
      \cmidrule(r){1-1} \cmidrule(l){2-2}
		DrugBank & advil, Benzylamine, Cobalt, S-Acetyl-Cysteine, Wellbutrin, zzzquil\\
		MeSH &Anaphylaxis, Cough, Drainage, Hospitalization, Neoplasms, Self-Testing\\
		Manual & \#antivaxxer, \#cancersucks, \#depressionisreal, \#mswarrior, \#plantbasedhealing, \#SocialDistancing\\
     \bottomrule
\end{tabularx}
\caption{Example terms from each sampling method.}
 \label{tab:search-term-examples}
\end{table}

\subsection*{Additional Examples from the Corpus}
Table \ref{tab:shortest-longest-tweet} shows the shortest and longest tweets in the dataset.

\begin{table}[h!]
\centering\small
\renewcommand\tabularxcolumn[1]{m{#1}}

\begin{tabularx}{\columnwidth}{lrX}
\toprule
id & \#words & tweet \\
\cmidrule(lr){1-1} \cmidrule(lr){2-2} \cmidrule(lr){3-3}
1 & 4 & bpd symptoms on 1000\\
\cmidrule(lr){1-3} 
2 & 4 & Increasing pain unlocked \#PTSDAwarenessDay\\
\cmidrule(lr){1-3}
3 & 114 & @username [...] @username I've been to every hospital in my region, "Sorry, can't help you" I don't want drugs, I want my back fixed. I know for a fact the tech is there. They don\'t want the liability. They should just Quit Medicine! I'm called inoperable with intractable pain, none will help. No Pain RX \\

\bottomrule

\end{tabularx}
\caption{Longest and shortest tweet in the dataset.}
\label{tab:shortest-longest-tweet}
\end{table}

\subsection*{Annotation Aggregation Strategies}
\label{appendix:aggregation}

We provide an aggregated version of the dataset which adjudicates both annotators' results.
In general, our strategy is motivated by a high recall approach to ensure we do not lose any annotated perspectives on the data. When combining the annotations, we choose the longest overlapping sequence between two instances. We prefer more frequent entity and relation classes over less frequent ones, and choose more general concepts over more specific ones.
We aggregate in two steps by first aligning the entity annotations, followed by aggregating the relations.
\paragraph{Entities} With regards to the entity span, we use the longest overlapping span between A1's and A2's annotation. In cases in which they disagree on the entity type, we chose the more frequent class. Exceptions are the entity classes \textit{treatment} and \textit{biochem}. For those classes, one subgroup is more general than the other. If both annotators agree on the major class (\textit{treat}), but disagree on the subtype (\textit{drug} vs. \textit{therapy}) we aggregate to the more general one which are \textit{treat\_therapy} or \textit{biochem\_substance}. 

For cases in which one annotator labeled an entity as \textit{other} while the second annotator chose a different entity class, we aggregate to the more frequent entity class. However, if the annotator used \textit{other} to model a relation, we keep the entity as \textit{other} to keep the relation intact and valid.
\footnote{In the guidelines annotators are instructed to prioritize assigning an accurate  relation over an accurate entity type. In some cases this means they may default to an \textit{other} entity if the relation they want to model is not allowed for a particular entity pair.}

If one annotator labeled an entity, but the other one did not, we generally follow a high recall approach and add this entity to the aggregated document. However, we additionally check if the annotator who marked the entity used it to model a relation. If the relation is valid (i.e. the involved entities are allowed to be connected), we use the entity, otherwise it is dropped.
\paragraph{Relations} 
To adjudicate the relation annotation, we identify cases in which both annotators agreed on the fact that there is any type of relation between a given entity pair. First, we check if the relation tags are valid (i.e. the involved entities are allowed to be connected). If one of them is invalid, we choose the valid one for the aggregated version. If both are invalid, the relation is dropped. If they are both valid, we choose the more frequent relation class.
One exception to this rule concerns cases in which one annotator identified an \textit{other} relation while the second annotator chose a different relation class. Here, the tag  \textit{other} indicates a vague relation which is not in line with our aim to adjudicate to the more specific class. Therefore, we can not resolve this by simply assigning the more frequent label, because some of the small relation classes are less frequent than the class  \textit{other}. A1 and A2 consequently revisit those cases (11 instances) and decide jointly which relation type should be added to the aggregated version.

For annotations in which A1 and A2 only agreed on one of the involved entities, 
we follow a high recall approach and keep both relations for the adjudicated version of the data as long as the relations are valid.
Finally, we consider cases in which one annotator did not label any relation while the other identified one. For those, we hypothesize that they are ambiguous and that the missing relation reflects that (i.e. that the relation marked by one of the annotators might be covering a political claim about a medical topic). In an effort not to lose these borderline cases, we add them to the aggregation as long as the relation is valid.

\end{document}